\documentclass[conference]{IEEEtran}
\IEEEoverridecommandlockouts
\usepackage[backend=biber,style=ieee]{biblatex}
\usepackage{amsmath,amssymb,amsfonts}
\usepackage{algorithmic}
\usepackage{graphicx}
\usepackage{subcaption}
\usepackage{textcomp}
\usepackage{xcolor}

\def\BibTeX{{\rm B\kern-.05em{\sc i\kern-.025em b}\kern-.08em
    T\kern-.1667em\lower.7ex\hbox{E}\kern-.125emX}}

\begin{document}

\title{A Deep Convolutional Neural Network-based Model for Aspect and Polarity Classification in Hausa Movie Reviews}

\author{
\IEEEauthorblockN{1\textsuperscript{st} Umar Ibrahim}
\IEEEauthorblockA{\textit{Dept. of Computer Science} \\
\textit{Federal University Dutse}\\
Jigawa, Nigeria \\
faruuk377@gmail.com}
\and
\IEEEauthorblockN{2\textsuperscript{nd} Abubakar Yakubu Zandam}
\IEEEauthorblockA{\textit{Dept. of Computer Science} \\
\textit{Federal University Dutse}\\
Jigawa, Nigeria \\
ayzandam95@gmail.com}
\and
\IEEEauthorblockN{3\textsuperscript{rd} Fatima Muhammad Adam}
\IEEEauthorblockA{\textit{Dept. of Computer Science} \\
\textit{Federal University Dutse}\\
Jigawa, Nigeria \\
fateemamohdadam2@gmail.com}
\and
\IEEEauthorblockN{4\textsuperscript{th} Aminu Musa}
\IEEEauthorblockA{\textit{Dept. of Computer Science} \\
\textit{Federal University Dutse}\\
Jigawa, Nigeria \\
musa.aminu@fud.edu.ng}
}

\maketitle

\begin{abstract}
Aspect-based Sentiment Analysis (ABSA) plays a pivotal role in understanding the nuances of sentiment expressed in text, particularly in the context of diverse languages and cultures. This paper presents a novel Deep Convolutional Neural Network (CNN)-based model tailored for aspect and polarity classification in Hausa movie reviews, an underrepresented language with limited resources and presence in sentiment analysis research. One of the primary implications of this work is the creation of a comprehensive Hausa ABSA dataset, which addresses a significant gap in the availability of resources for sentiment analysis in underrepresented languages. This dataset fosters a more inclusive sentiment analysis landscape and advances research in languages with limited resources. The collected dataset was first preprocessed using sci-kit-learn to perform TF-IDF transformation for extracting feature word vector weights. Aspect-level feature ontology words within the analyzed text were derived, and the ontology of the reviewed text was manually annotated, along with corresponding sentiment analysis polarity assignments. The proposed model was developed by combining convolutional neural networks (CNNs) with an attention mechanism to aid in aspect-word prediction. The model utilizes sentences from the corpus and feature words as vector inputs to enhance prediction accuracy. The proposed model leverages the advantages of the convolutional and attention layers to extract contextual information and sentiment polarities from Hausa movie reviews. Its excellent performance demonstrates the applicability of such models to underrepresented languages. With 91\% accuracy on aspect term extraction and 92\% on sentiment polarity classification, the model excels in aspect identification and sentiment analysis, offering insights into specific aspects of interest and their associated sentiments. Moreover, the proposed model outperformed traditional machine models in both aspect-word and polarity prediction. This study, through the creation of the Hausa ABSA dataset and the development of an effective model, significantly advances ABSA research and has wide-ranging implications for the sentiment analysis field in the context of underrepresented languages and cross-cultural linguistic research.
\end{abstract}


\section{Introduction}
%This document is a model and instructions for \LaTeX.
%Please observe the conference page limits. 

In the realm of natural language processing, our study delves into the realm of aspect-based sentiment analysis within the context of the Hausa language—a domain characterized by scarce linguistic resources. Our core aim is to harness the potential of Convolutional Neural Networks (CNNs) as formidable tools to dissect sentimental nuances\cite{dang2021hybrid}. Recent work serves as an example of this pursuit. Despite the constraints imposed by scarce linguistic resources, our endeavor is driven by the desire to enhance sentiment analysis accuracy. Through this, we seek to illuminate the intricate interplay of sentiments across diverse dimensions. Our goal is to unravel the dynamics of sentiment within this resource-constrained language, ultimately deepening our understanding of Hausa's distinctive linguistic and cultural context.

In today's digital age, users of movie streaming platforms like Amazon Movies, Netflix, and YouTube have a plethora of means to express their opinions via comments and rating systems. These opinions encapsulate a spectrum of sentiments ranging from positive to negative. Businesses leverage data mining techniques to comprehend these sentiments, thereby improving customer experiences and catering to their preferences. However, the inherent complexities arise from factors like misspelled words, abbreviations, emoticons, and slang. Consequently, selecting an optimal opinion mining method becomes pivotal in generating precise user sentiment \cite{onalaja2021aspect}.

Sentiment analysis (SA) constitutes a computational approach to discerning and categorizing opinions expressed in text, intending to determine whether the writer's opinion about a particular subject or product is positive, negative, or neutral. Within the broader context of behavioral science, SA serves as a fundamental tool for interpreting and classifying emotions across diverse domains \cite{sanisentiment}. This branch of analysis delves into people's sentiments, beliefs, views, and feelings regarding products and services, topics, events, and their respective attributes \cite{ullah2022deep}.

While existing studies have examined sentiment analysis in Hausa, a low-resource language, with a primary focus on sentiment analysis within an entire document \cite{sanisentiment,muhammad2022naijasenti,abubakar2021enhanced,zandam2023online,rakhmanov2022sentiment}, this research introduces a novel dimension by venturing into Aspect-Based Sentiment Analysis (ABSA) using a dataset generated from Hausa YouTube comments. Our pioneering approach aims to provide sentiment analysis at the word level, forging new pathways in this field of inquiry.

Inspired by the swift progress of deep learning and the effectiveness of CNNs in text analysis. Because CNN  can capture local features, word sequences, and hierarchical structures. our innovative approach seeks to improve traditional sentiment analysis by operating at the word level, thus opening novel directions in this research domain.

The rest of the paper sections are organized as follows: Section II examines relevant studies, Section III describes the research methodology employed, Section IV presents the significant results along with a discussion of the findings, and Section V concludes the study, offering direction for future research

\subsection {Background }

Aspect-based sentiment analysis (ABSA) involves the task of discerning the sentiment associated with a particular term within a text. It can be regarded as an advanced form of sentiment analysis, where the key aim is to recognize specific aspect terms within the provided reviews and anticipate the sentiments linked with those distinct aspects. The initial step in conducting ABSA is known as Aspect Term Extraction (ATE), wherein the phrases that represent the core subjects of opinions within a review are identified \cite{li2018aspect}. To illustrate, consider the review: "I appreciate the scene and the producer's choice of soundtrack." In this case, the terms "scene" and "producer" will be extracted as aspect terms, and their sentiment will be determined by the presence of the opinion word "appreciate". This process constitutes the second subtask of ABSA, known as Aspect Term Sentiment Classification (ATSC).

ATE task was considered by some researchers to be a sequential modeling problem where a sentence is modeled as a set of tokens. Various algorithms, ranging from Support Vector Machines (SVM) \cite{castellucci2014unitor,akhtar2017feature} for ATE to Attention-based LSTM  for ATSC\cite{feng2022unrestricted}, have been proposed to facilitate ABSA's accurate execution.

A considerable range of methods for sentiment analysis across different domains already exists \cite{go2009twitter,saif2012semantic,krishna2019sentiment}. The majority of these existing studies mainly concentrate on identifying sentiments at the textual level\cite{rhanoui2019cnn} or in a single sentence \cite{basiri2017sentence}. However, these approaches often fall short of providing the specific and detailed insights users seek from reviews, particularly regarding individual aspects, features, or attributes of a product or service. For example, consider a user review “I enjoyed the movie, but the character was annoying”. This demonstrates how viewers can have varied perspectives, where their focus might be on the quality of the movie while at the same time being annoyed by the actor.  Hence, it becomes pivotal to gauge sentiments about distinct aspects, a task referred to as aspect-based analysis. 

ABSA holds the promise of transforming sentiment analysis from a broad perspective to a granular level, offering a more specific understanding of opinions. By analyzing sentiments at the aspect or feature level, ABSA empowers businesses to gain actionable insights from customer feedback. This fine-grained approach not only aids in pinpointing areas of improvement but also enhances market research, competitive analysis, and domain-specific sentiment understanding. ABSA's potential extends beyond industries; it facilitates personalized recommendations, guides product development, and enables timely responses to evolving sentiment trends.
 The insights drawn from ABSA support evidence-based decision-making in policy and regulations, ensuring that public sentiments are understood and considered. Making it an important tool for understanding and harnessing sentiments in diverse contexts.

One of the major challenges of ABSA in low-resource languages is the absence of rich and diverse datasets. Additional challenges include developing a high-quality labeled dataset for training the ABSA model and the scarcity of domain-specific annotated data. Extending ABSA-trained models to different languages, as in the case of multilingual models, also presents difficulties in handling linguistic variations, idiomatic expressions, and sentiment lexicons across diverse languages. Therefore, collecting datasets for each language and training the ABSA model on language-specific datasets is vital to achieving accurate results.

ABSA is a subtask of sentiment analysis that identifies the sentiment of a sentence or phrase at the aspect level. This is important for low-resource languages like Hausa because it allows us to gain insights into the sentiment of people about specific aspects of a product or service, even when there is limited data available.
%%On the other hand, CNNs are a type of deep learning model that can be used for ABSA. CNNs are well-suited for this task because they are able to learn local patterns in text \cite{dang2021hybrid}, which is important for identifying aspects and their corresponding sentiment. CNNs are able to learn more generalizable features from limited data, which is important for low-resource languages.

\section{LITERATURE REVIEW}
This literature review aims to provide an overview of the current state-of-the-art methods and recent developments in Aspect-Based Sentiment Analysis(ABSA) techniques. ABSA evolved naturally as an extension of sentiment analysis to address the need for a more fine-grained analysis of sentiments concerning specific aspects or features. Although many researchers contributed to the development of ABSA, each tried to address a drawback of traditional SA.

 ABSA is a specialized area of sentiment analysis that focuses on extracting sentiment information at the aspect level from text \cite{wang2016attention}. ABSA has been an improvement over the traditional method of sentiment analysis, in which texts are classified as positive or negative only without considering contextual information\cite{socher2013recursive}. ABSA  can be formulated in different ways and techniques;\cite{sebastiani2006sentiwordnet} conducted a thorough survey of ABSA highlighting challenges and methodologies in aspect-based sentiment analysis. The authors further proposed a technique for identifying aspects and associating sentiments.
 
 \cite{tang2016aspect} proposed a technique called the rating regression approach, where aspects are identified and rated by a given score. The mathematical model clearly outlines how ABSA can be further developed. Ever since, more efforts have been made to build an aspect-based analysis system.
 
 Baccianella \cite{xu2019bert} introduced SentiWordNet, a lexical resource that assigns sentiment scores to words based on their synsets. SentiWordNet has been widely used as a feature in traditional machine learning models for aspect-based sentiment analysis to determine sentiment polarity based on the words associated with aspects.
 
Additionally, Poalo \cite{xue2018aspect} demonstrates the use of SentiWordNet in sentiment analysis with Twitter data, with a focus on determining sentiment orientation at the document level. This concept was extended by Brychc \cite{tang2015effective}  to aspect-level analysis, by associating SentiWordNet scores with aspects and utilizing traditional machine learning models to classify the sentiment of individual aspects.

Recently, the popularity of deep learning has led to the integration of neural networks in many fields such as business and healthcare\cite{8949669,9803131}. Also in ABSA deep learning has brought about substantial advancements.  \cite{wang2018aspect,he2018exploiting,ni2018learning,wu2021context} all proposed deep-learning-based ABSA techniques using high-resource language datasets,  leading to more accurate and nuanced sentiment analysis models \cite{xu2019bert}. For example, \cite{xue2018aspect} introduced deep learning-based ABSA by proposing an attention-based LSTM model for aspect-level sentiment classification. The authors proposed a hierarchical model that first captures the context of the whole sentence and then focuses on individual aspects using attention mechanisms. The model effectively addresses the challenge of aspect-level sentiment classification by incorporating both local and global context information. 

Furthermore,  \cite{rosso2011exploiting} delved into an end-to-end approach of combining aspect extraction and aspect sentiment classification, which they considered more advantageous than considering the tasks as separate entities. In the past, when working on ABSA, people didn't use sentence structure clues well. This made it hard for the aspect extraction to find multi-word aspects, and the aspect sentiment classifier couldn't understand how words fit together in a sentence. The proposed solutions demonstrate superior performance compared to existing models on the SemEval-2014 dataset, excelling in aspect extraction and aspect sentiment classification subtasks.

More so, \cite{thet2010aspect} leveraged movie reviews on discussion boards to explore aspect-based sentiment analysis. This approach performed sentiment analysis at the clause level to ensure that different aspects were analyzed separately.

Consequently, \cite{essebbar2021aspect} acknowledged the existence of many ABSA works in English using Pre-Trained Models (PTM) and also highlighted the availability of open datasets. The author stressed that ABSA works in French are few as compared to many SA works. This prompted the author to utilize the French Language and three fine-tuning methods such as French PTM multilingual BERT (mBERT), CamemBERT, and FlauBERT on the SemEval2016 datasets. The work surpassed the state-of-the-art French ABSA with flexibility for out-of-domain Datasets. In line with low-resource languages, \cite{bensoltane2022towards} proposed a transfer learning-based approach for Arabic ABSA.

To the best of our knowledge, there have not been many works on aspect-based sentiment analysis in low-resource languages. Therefore, In this work, we tried to make the first attempt at organizing the first Hausa language dataset for ABSA and propose a deep learning model for Hausa Aspect-based sentiment analysis.

\section{METHODOLOGY}

\subsection{Dataset Description:}
The dataset is generated from various YouTube channels where comments were made by individuals about Hausa movies. Within this dataset, comments were categorized as either purely Hausa or Engausa\footnote{
 a mixture of Hausa and English
 }.The dataset comprises 590 comments, with four features consisting of comments, aspect words (1-person [actor, actress, director], 2-episode, 3-Movie, 4-general), sentiment polarity (negative or positive), and language (Hausa, Engausa).
 
\subsection{Dataset collection and Preprocessing}
 Dataset collection was carried out using web scraping with Python, which facilitated the gathering of numerous  Hausa comments from YouTube channels. Subsequently, the dataset undergoes preprocessing, involving the removal of instances that did not capture any of the aspect categories and the removal of stopwords—words that typically contribute less meaning to a text, such as 'a,' 'ni,' 'to,' 'su,'an'. Additionally, irrelevant terms, such as emojis, usernames, special characters (@, \#), punctuation marks, and numbers, were also removed. Furthermore, the texts were converted to lowercase. Finally, comments and aspect words were vectorized using Term Frequency Inverse Document Frequency (TF-IDF), one of the most straightforward and effective ways to comprehend the context of a document and locate relevant information and key terms and phrases.
 
\subsection{Aspect Term Extraction Processing} 
From the dataset, the aspect terms were manually annotated by native speakers, looking at both the context and semantic meaning of the review. Some predefined rules were followed for aspect-term annotation. For example, all proper nouns represent either a person or a movie aspect category. We then applied a supervised learning method for aspect-term extraction, specifically, a deep convolutional neural network was used with an attention layer. To enable implicit aspect extraction and understanding, 
We split the datasets for each task into a training set with 70\% of the comments and a test set with the remaining comments. A hyperparameter tuning strategy was utilized to find the best parameter sets.

\subsection{Model building}
 
This section presents the proposed model used in this paper. The model is a deep convolutional neural network with recurrent and attention layers. The recurrent layers allow the model to scan from left to right to identify the semantics of the sentence, while the attention layer allows the model to focus on specific important words in the sentence before the aspect term is extracted. Fig. 1 provides a clear description of the entire model framework.

\begin{figure}[htpb]
\centering
        \includegraphics[width=0.8\linewidth]{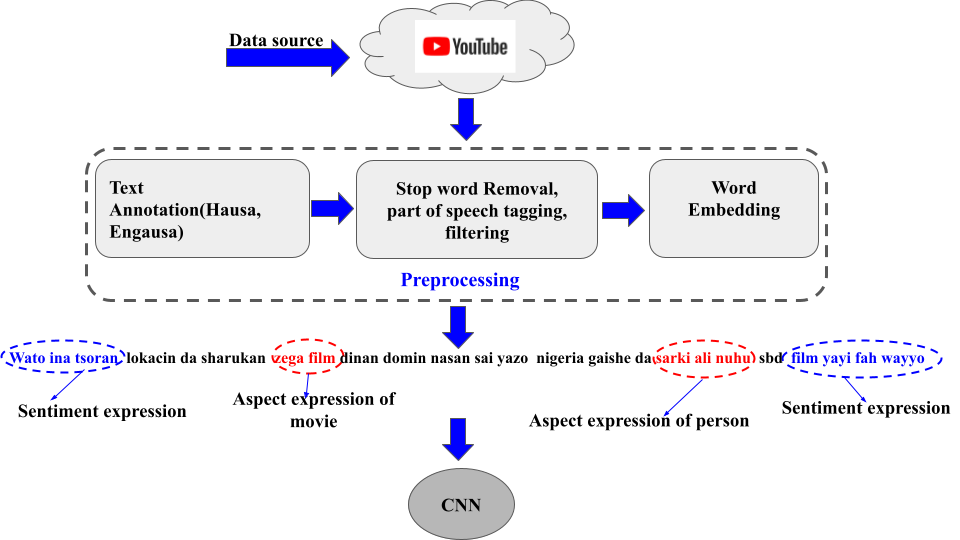}
        \caption{Proposed methodology}
        
\end{figure}

The proposed CNN Model architecture can be described as a 9-layer DCNN. The layers are arranged as follows: The first layer is an embedding layer that converts the input data into a dense vector representation. This layer is used to learn the embeddings of the input words, which are then used as input for the rest of the layers. Then there are two Conv1D layers, which are the convolutional layers of the model. These layers are used to learn feature maps from the input data. The first convolutional layer has 32 filters of size 2, and the second layer has 64 filters of size 2. The next layer is the LSTM layer, a recurrent layer that allows the model to pass information from left to right in the sequence. An attention layer is added, which takes embedding from the LSTM layer as input. A GlobalMaxPooling1D layer is added to reduce the spatial dimension of the feature maps learned by the attention layer. The layer is followed by a dense layer with 256 neurons, which is used to learn higher-level features from the feature maps produced by the previous layers. The next layer is a dropout layer, which is used to reduce overfitting by randomly dropping out some of the neurons during training. Finally, a dense layer with 4 units and softmax activation is added as the output layer. The proposed model architecture is presented in Fig. 2 to highlight the layers and available parameters.

\begin{figure}[htpb]
\centering
        \includegraphics[width=0.8\linewidth]{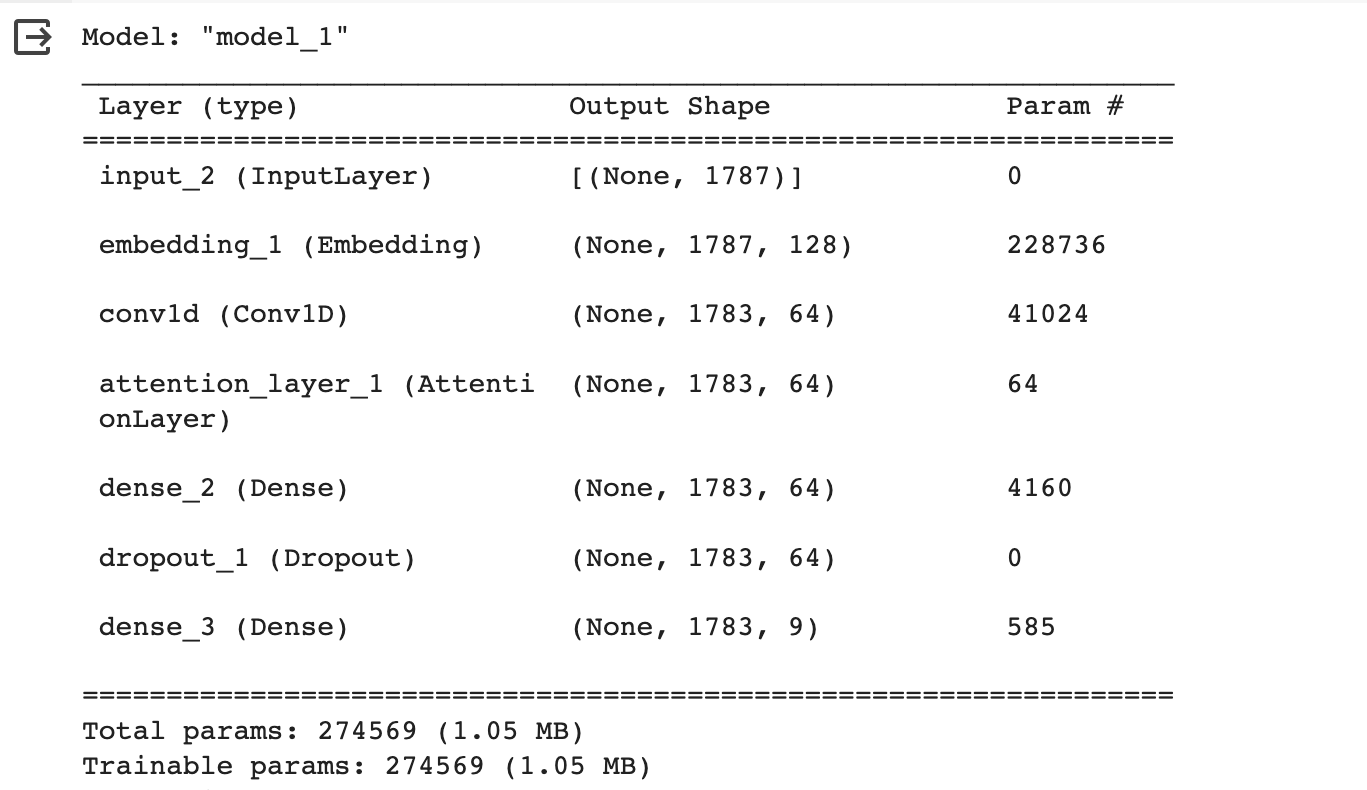}
        \caption{Model architecture}
        
\end{figure}

\subsection{Performance metrics:} 
To assess the performance or effectiveness of a model, system, or process, we employ performance metrics, which are quantitative measures.
\begin{itemize}
    \item Accuracy: The number of successfully categorized instances (aspect-based sentiment) divided by the total number of cases (aspect-based sentiment instances). It can be calculated using the following formula:
    \[
    \text{Accuracy} = \frac{\text{True Positives} + \text{True Negatives}}{\text{Total Instances}}
    \]
    
    \item Precision: is calculated as the proportion of true positive instances (positive sentiment polarity) correctly classified as positive. The formula for precision is as follows:
    \[
    \text{Precision} = \frac{\text{True Positives}}{\text{True Positives} + \text{False Positives}}
    \]

    \item Recall: also known as sensitivity, is the proportion of positive instances correctly classified as positive. The formula for recall is given by:
    \[
    \text{Recall} = \frac{\text{True Positives}}{\text{True Positives} + \text{False Negatives}}
    \]

    \item F1 Score: The F1 score is a metric that combines both precision and recall, striking a balance between them. It can be computed using the following formula:
    \[
    \text{F1 Score} = \frac{2 \cdot \text{Precision} \cdot \text{Recall}}{\text{Precision} + \text{Recall}}
    \]
\end{itemize}

\section{ Result and Discussion} 

\subsection{Preliminary analysis} 
A descriptive analysis conducted on the dataset revealed that the person aspect term appeared the most, while in terms of sentiment polarity, neutral and positive are the majority classes with a few negative instances. The results of the analysis are presented in Fig. 3.

\begin{figure}[htbp]
    \centering
    \begin{subfigure}{0.5\linewidth}
        \centering
        \includegraphics[width=\linewidth]{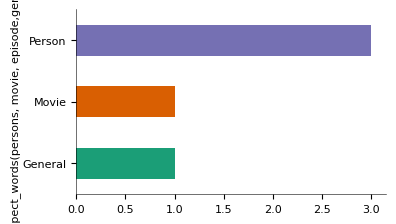}
        \caption{}
        \label{fig:aspect}
    \end{subfigure}%
    \begin{subfigure}{0.5\linewidth}
        \centering
        \includegraphics[width=\linewidth]{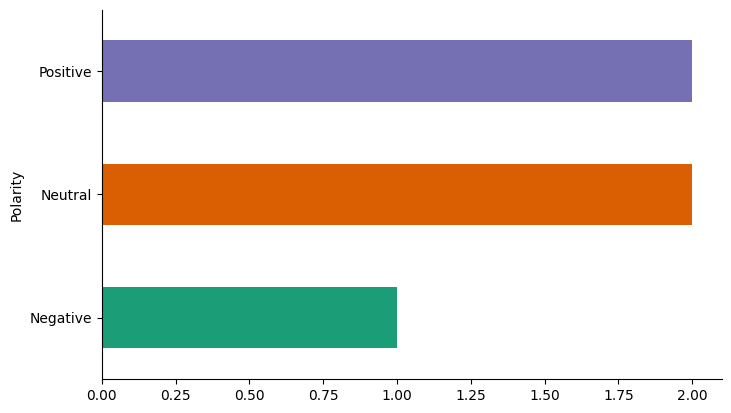}
        \caption{}
        \label{fig:polarity}
    \end{subfigure}
    \caption{Distribution of aspect terms and sentiment polarity in the dataset}
    \label{fig:distribution}
\end{figure}

The plots indicate that the aspect-based sentiment from the comments we generated shows that viewers are more interested in the person aspect words, which can be either "director", "actor", or "actress".

\subsection{Performance of the proposed model}
The performance of the proposed deep learning-based aspect-level sentiment analysis is presented in Fig. 4 and 5 The model was trained on two separate tasks: identifying aspect words and polarity classification of the aspect words.  The model was also evaluated on both aspect word extraction and aspect word polarity classification.

\begin{figure}[htbp]
  \centering
  \begin{subfigure}[b]{0.5\linewidth}
    \includegraphics[width=\linewidth]{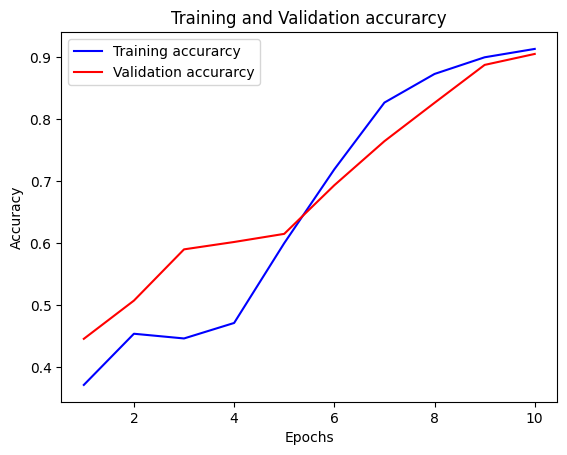}
    \caption{Aspect Accuracy}
    \label{fig:aspect_accuracy}
  \end{subfigure}%
  \begin{subfigure}[b]{0.5\linewidth}
    \includegraphics[width=\linewidth]{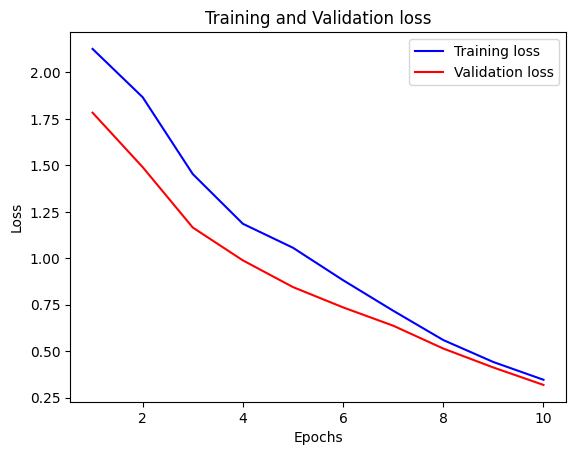}
    \caption{Aspect Loss}
    \label{fig:aspect_loss}
  \end{subfigure}
  \caption{Accuracy and loss curves of aspect word extraction}
  \label{fig:aspect_curves}
\end{figure}

In Fig. 3, the performance of the proposed deep learning model is evaluated with accuracy and loss curves. the results indicate superior performance over the baseline machine learning model in identifying aspect words from a given review. The model performance on aspect word polarity classification is presented in Fig. 4.

\begin{figure}[htbp]
  \centering
  \begin{subfigure}[b]{0.5\linewidth}
    \includegraphics[width=\linewidth]{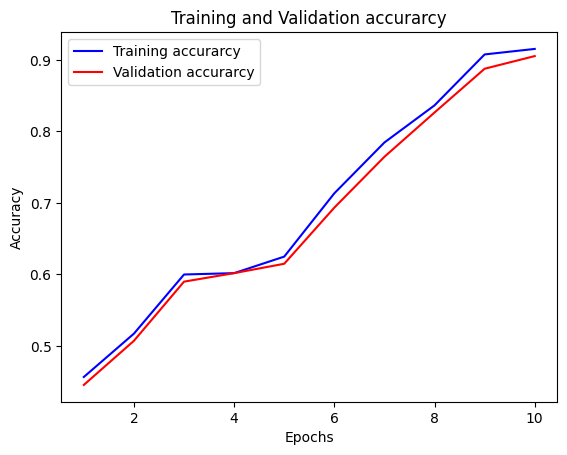}
    \caption{Polarity Accuracy}
    \label{fig:polarity_accuracy}
  \end{subfigure}%
  \begin{subfigure}[b]{0.5\linewidth}
    \includegraphics[width=\linewidth]{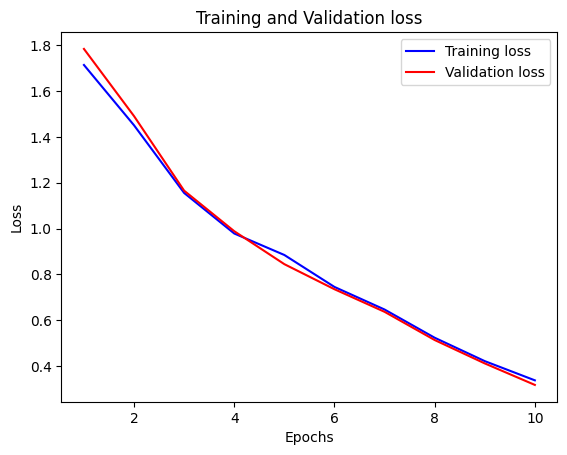}
    \caption{Polarity Loss}
    \label{fig:polarity_loss}
  \end{subfigure}
  \caption{Accuracy and loss curves of sentiment polarity classification}
  \label{fig:polarity_curves}
\end{figure}

The proposed model achieved remarkable accuracy in aspect word polarity classification of 92\%. Compared to baseline models, the proposed model 
achieved the best results in terms of accuracy in all the two tasks. This was expected because a Deep neural network with an attention mechanism is the most appropriate for tasks that involve sequential data, which is the case for the extraction of implicit aspect words. Traditional machine learning performs poorly in aspect word detection and polarity classification because the models work better on numerical tabular data. The baseline model results for aspect word detection and polarity classification are presented in Table I and II, respectively.
\begin{table}[ht]
\centering
\caption{Performance of traditional machine learning of aspect word Extraction}
\begin{tabular}{lcccc}
\hline
Model & Accuracy & Precision & Recall & F1-score \\
\hline
Naive Bayes & 0.70 & 0.71 & 0.70 & 0.67 \\
SVM & 0.7153 & 0.6900 & 0.7211 & 0.7021 \\
Random Forest & 0.7000 & 0.72 & 0.70 & 0.70 \\
Logistic Regression & 0.7624 & 0.7411 & 0.7602 & 0.7511 \\
\hline
\end{tabular}

\label{tab:aspect_word}
\end{table}

\begin{table}[ht]
\centering
\caption{Performance of traditional machine learning on aspect polarity }
\begin{tabular}{lcccc}
\hline
Model & Accuracy & Precision & Recall & F1-score \\
\hline
Naive Bayes & 0.64 & 0.60 & 0.64 & 0.52 \\
SVM & 0.64 & 0.60 & 0.64 & 0.52 \\
Random Forest & 0.64 & 0.60 & 0.64 & 0.52 \\
Logistic Regression & 0.64 & 0.60 & 0.64 & 0.52 \\
\hline
\end{tabular}

\label{tab:model_polarity}
\end{table}
From Tables I and II, the machine learning methods of logistic regression achieved a result that is superior to the rest of the baseline models in both aspect term detection and classification, while Naive Bayes (NB), Random Forest, and SVM had similar results. The proposed model's accuracy is compared with the baseline models in Table III. The results show that the proposed models outperformed baseline models in terms of accuracy.

\begin{table}[ht]
\centering
\caption{Performance comparison between proposed model and machine learning on aspect and polarity task}
\begin{tabular}{l c}
\hline
Model & Accuracy \\
\hline
\multicolumn{2}{l}{Result on Aspects Extraction:} \\
Naive Bayes & 0.64 \\
SVM & 0.64 \\
Random Forest & 0.64 \\
Logistic Regression & 0.64 \\
Proposed DCNN model & 0.91 \\
\hline
\multicolumn{2}{l}{Result on Polarity classification:} \\
\hline
Naive Bayes & 0.64 \\
SVM & 0.64 \\
Random Forest & 0.64 \\
Logistic Regression & 0.64 \\
Proposed DCNN model & 0.92 \\
\hline
\end{tabular}
\label{tab:model_polarity_2}
\end{table}

\section{Conclusion and Future work}
In this research work, we presented a deep learning-based aspect-level sentiment classification of movie reviews in the Hausa language.
The main contributions of this work center on providing a Hausa movie review dataset for aspect-level sentiment analysis. We also proposed a CNN model that learns aspect embeddings in Hausa textual data, a language categorized as underrepresented in natural language processing regarding resources. This provides a starting point for aspect-based sentiment analysis in the Hausa language. The proposed model achieves remarkable performance on both aspect-term extraction and polarity classification, despite the small size of the dataset. The attention layer allows the model to concentrate on a different part of the sentence to discover aspect terms, allowing it to accurately perform aspect-polarity classification. The proposed model outperforms all the compared baseline models which used traditional machine learning models with significant differences.

The major challenge of this work is the inability of the model to analyze more than one aspect word in a sentence at a time. Therefore, an intriguing and potential avenue for future work would be to investigate the potential of transformer architecture in aspect-based sentiment analysis to handle the aforementioned challenge of extracting and analyzing all aspect terms in a sentence simultaneously.

\section*{Acknowledgment}

This work is partially funded by an unrestricted gift from Google. Under Google exploreCSR 2023 award.

%\section*{References}

\printbibliography
\end{document}